\documentclass[10pt,twocolumn,letterpaper]{article}

\usepackage{algorithm}
\usepackage{algpseudocode}
\usepackage{pifont}
\usepackage{multirow}
\usepackage[pagenumbers]{wacv} 

\definecolor{wacvblue}{rgb}{0.21,0.49,0.74}
\usepackage[pagebackref,breaklinks,colorlinks,allcolors=wacvblue]{hyperref}

\def\wacvPaperID{3303} 
\def\confName{WACV}
\def\confYear{2027}

\title{Dynamic Time Step Prediction in Inverse Heat Dissipation for Blur-Like Image Restoration Tasks}

\author{%
  Cap Dang Xuan Kiet\\
    Nanyang Technological University\\
    Singapore\\
  \texttt{capd0001@e.ntu.edu.sg} \\
 \and
Tat-Jen Cham\\
Nanyang Technological University\\
Singapore\\
  \texttt{astjcham@ntu.edu.sg} \\
}

\begin{document}
\maketitle
\begin{abstract}
When using diffusion models to target image restoration problems, diffusion inversion is typically employed to retain relevant image information from the degraded images.
Instead of inverting back to the initial time step (\ie $T$), many methods invert to a pre-determined intermediate time step, in order to better preserve information from degraded source images. However, a pre-determined time step for inversion is not ideal for reconstruction, as a severely degraded image requires an earlier starting time step than a mildly degraded one. In addition, DDIM-based models corrupt the original signal by adding Gaussian noise, which can be mismatched to the nature of blur-like degradations, such as blur, haze, and low-light. 
To address these problems, we propose two solutions: (1) we adopt an alternative diffusion process, called the Inverse Heat Dissipation Model, that diffuses the input image by gradually blurring a  data point (2) we propose to implement a time predictor to estimate the starting time step for the inversion, with the model learning to adapt to the degradation severity. Extensive experiments on standard benchmarks show that our method achieves state-of-the-art performance in both quantitative and qualitative evaluations, with excellent generalization to many restoration tasks.
\end{abstract}
    
\section{Introduction}

\label{sec:intro}

Image restoration aims to recover clean images from degraded observations. Because of the information loss from the degradation process, the reverse problem can have multiple feasible solutions. This ill-posed nature makes image restoration a highly challenging problem. Many excellent restoration models have been proposed \citep{Chen2020PreTrainedIP,Dong2020MultiScaleBD,Liang2021SwinIRIR,Ren2019SingleID,Wang2021UformerAG,Zamir2020CycleISPRI,Zamir2021RestormerET}, but their generation abilities are limited. Recent advances in the diffusion model \citep{NEURIPS2020_4c5bcfec} have showcased its remarkable ability in high-quality image generation. 
Some prevailing papers \citep{Chung2021ComeCloserDiffuseFasterAC, Tumanyan2022PlugandPlayDF} suggest using the input image as a prior via the DDIM Inversion strategy \citep{Song2020DenoisingDI} when editing an image to secure its contextual and structural information. To this end, many algorithms usually attempt to modify the intermediate features of the diffusion process with the given degraded observation. 
\citep{Wang2022ZeroShotIR,Kawar2022DenoisingDR} are proposed to iteratively refine the reverse process, constraining diffusion priors with the given input through optimization. \citep{Lin2024DiffBIRTB,Wu2023SeeSRTS,Yang2023PixelAwareSD} follow ControlNet \citep{Zhang2023AddingCC} design, attempting to finetune diffusion priors with input image. 

The effectiveness of any inversion process critically depends on the choice of the initial state derived from the input image, as illustrated in Fig.~\ref{fig:3modelsdiff}(a). In DDIM-based methods, the forward diffusion process progressively injects Gaussian noise, mapping a degraded data point from the image manifold to the Gaussian manifold. While this formulation naturally aligns with degradations that are noise-like, it is less suitable for blur-like degradations, including motion blur, haze, and low-light effects. These degradations arise from structured physical processes rather than stochastic noise injection, leading to a mismatch between the degradation model and the forward diffusion dynamics. To address this issue, we propose replacing the standard noise-based diffusion with a diffusion-like iterative generation model that employs a blurring forward process. By modeling the forward trajectory using a blurring mechanism instead of random noise, the resulting intermediate states remain more physically aligned with the underlying degradation. This improves the predictability of the initial state for inversion and facilitates a more accurate reconstruction. Specifically, we adopt the Inverse Heat Dissipation Model (IHDM) \citep{rissanen2023generative}, in which the forward destruction process is governed by deterministic, noise-free heat dissipation across time steps. This formulation better reflects the nature of blur-related degradations and enables a more coherent inversion pathway.

\begin{figure*}[t]
    \centering
    \captionsetup{type=figure}
    \includegraphics[width=\textwidth]{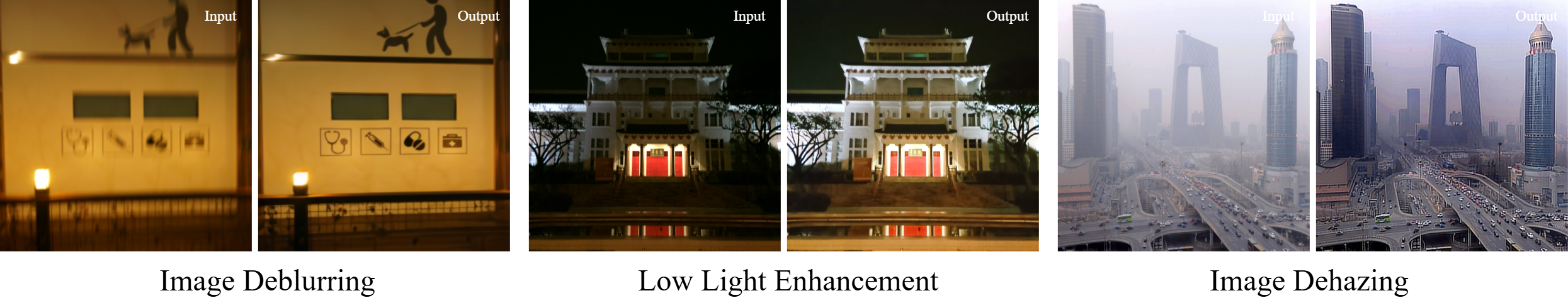}
    \captionof{figure}{Real-world applications of our method. We use images from the realistic dataset RealBlur \citep{Rim2020RealWorldBD} for the deblurring task and internet images for the dehazing and low-light enhancement tasks. Our method can generate high-fidelity results.}
    \label{fig:intro}
\end{figure*}

In \citep{Yue2024ArbitrarystepsIS}, the degraded image is inverted to a predefined intermediate time step rather than to the terminal diffusion step $T$ in the Gaussian space, as illustrated in Fig.~\ref{fig:3modelsdiff}(b). Selecting a midway initial time step constrains the reverse process with partially preserved information from the forward process, enabling the model to leverage both the robustness of generative sampling and the structural cues retained in the degraded input. However, because the starting time step is fixed, the injected noise level remains constant. This assumption is acceptable for super-resolution with a fixed scale factor, but limiting for restoration tasks such as deblurring or dehazing, where the required noise depends on the severity of degradation. To sufficiently obscure the corruption and align intermediate features, heavily degraded inputs demand greater noise (i.e., an earlier time step) than mildly degraded ones. Motivated by this, we introduce a Time Step Predictor that estimates an appropriate initial time step conditioned on the degraded image. Rather than enforcing perfect alignment of intermediate features, our approach seeks the point along the dissipation trajectory where feature similarity is maximized.

A further key insight is that the dissipation process acts as a distribution mapping: it gradually transforms the degraded distribution toward a flat image space through the forward process and then reverses it back to the clean distribution. Along this trajectory, there exists an optimal time step at which the forward source distribution and the backward target distribution sufficiently overlap; insufficient or excessive overlap, respectively, leads to under- or over-constrained initial inversion features due to inadequate information from the degraded distribution. Identifying this optimal time step requires a specialized objective, as naive distance-based losses degenerate to a minimum at the final diffusion time step, where all inversion features collapse to trivial flat images. To address this, we employ mutual information computed from the intermediate features, which quantifies the shared information between the two distributions and provides a principled criterion for selecting the most suitable time step.

Our main contributions are threefold:
\begin{itemize}
    \item We explore the use of IHDM as the base model for the inversion process.
    \item We propose to estimate a precise initial time step for the dissipation inversion process, leading to a model that can well handle a huge range of critical levels of degradation. 
    \item To better evaluate the time step, we propose measuring the intermediate features alignment based on the overlap of the corresponding distributions.
    
\end{itemize}

\section{Related Works}
\label{sec:relatedworks}
\subsection{Multi-degradation Image Restoration}
While single degradation image restoration have received significant interest \citep{ding2020iqa,Dong2020MultiScaleBD,Zamir2020CycleISPRI,Ren2021AdaptiveCP,Zhang2017LearningDC,Song2022VisionTF,Zhang2020DeblurringBR,Nah2021CleanIA,Tsai2021BANetAB}, multi-degradation restoration is relatively under-explored in the field. \citep{Liu2022TAPETP,Valanarasu2021TransWeatherTR} focus on degradation caused by multiple weather conditions. They attempt to train an encoder or decoder to adapt to each degradation type. \citep{Li2020AllIO} proposes a unified framework for denoising, deraining, and dehazing, employing an image encoder trained via contrastive learning to capture effective representations of image degradation. These representations are subsequently used in a secondary network responsible for image restoration. The approach involves two-stage training, and its success largely depends on the accurate selection of positive and negative pairs as well as the quantity and diversity of training data. 
\citep{potlapalli2023promptir,tian2025degradation} proposes a drop-in module to learn the representation of multiple degradations, developing a single-stage training pipeline for all-in-one image restoration model.
\subsection{Diffusion Prior for Image Restoration} 
Diffusion prior is used in image restoration, following two main strategies: iterative re-optimization and fine-tuning. The former \citep{Chung2022ImprovingDM,chung2023diffusion,fei2023generative,murata2023gibbsddrm,song2022pigdm,xiao2024dreamclean,Yue2022DifFaceBF,Kawar2022DenoisingDR,Chung2021ComeCloserDiffuseFasterAC,Wang2022ZeroShotIR} attempts to align the intermediate results of the diffusion model with the given degraded input by solving optimization at every diffusion step. \citep{murata2023gibbsddrm,fei2023generative,Yue2022DifFaceBF} propose to use degradation models to facilitate the restoration via the reverse process. The technique compromises the model's performance in real-world scenarios. \citep{xiao2024dreamclean} guides the degraded input through the sampling phase by using its own diffusion prior as the initial pivot and preserving it throughout the sampling, allowing a faithful solution without knowing the degradation model. However, re-optimizing at every step of the diffusion process is greatly resource-consuming, leading to slow inference.

The latter \citep{Kim2024DiffusionMC,Nguyen2023VisualII,Wu2023SeeSRTS,Yang2023PixelAwareSD,Zhang2024DegradationGuidedOI,Li2024DistillationFreeOD,Xia2023DiffIRED,Yu2024ScalingUT} aims to fine-tune the pre-trained large diffusion model for the downstream task. The early methods \citep{Kim2024DiffusionMC, Lin2024DiffBIRTB} only rely on input images as control signals, overlooking the role of semantic text information in model pre-training. \citep{Wu2023SeeSRTS} extract degradation-sensitive semantic prompts to precisely control the pre-trained text-to-image model. However, it encounters difficulties when dealing with scenes containing a variety of objects or severely degraded images.

\subsection{Diffusion Inversion for Image Restoration} 
Diffusion inversion attempts to estimate the optimal noise set that, when processed through the diffusion model, reconstructs a given image. DDIM \citep{Song2020DenoisingDI} first addressed this by proposing a deterministic non-Markovian generation process for the diffusion model. \citep{rissanen2023generative} replaces random noise of the forward process with the heat equation, forming a deterministic process. To this end, subsequent approaches leverage either textual guidance \citep{Gal2022AnII, mokady2022nulltextinversioneditingreal}, or textual-and-visual guidance \citep{Miyake2023NegativePromptIF,Nguyen2023VisualII} to better optimize the model. \citep{Garibi2024ReNoiseRI,Ju2024PnPIB,Kang2024EtaID,Wallace2022EDICTED,Zhang2023ExactDI} refines intermediate noise map, leading to notable enhancements in inversion quality. 

To further improve image fidelity and adapt it to image restoration tasks, \citep{Chihaoui2024BlindIR} solves optimization at every inversion step. \citep{Yue2024ArbitrarystepsIS} introduces a noise prediction module that, once trained, enables efficient inversion without requiring iterative optimization during inference.

\section{Methodology}

\begin{figure*}
  \centering

    \includegraphics[width=\linewidth]{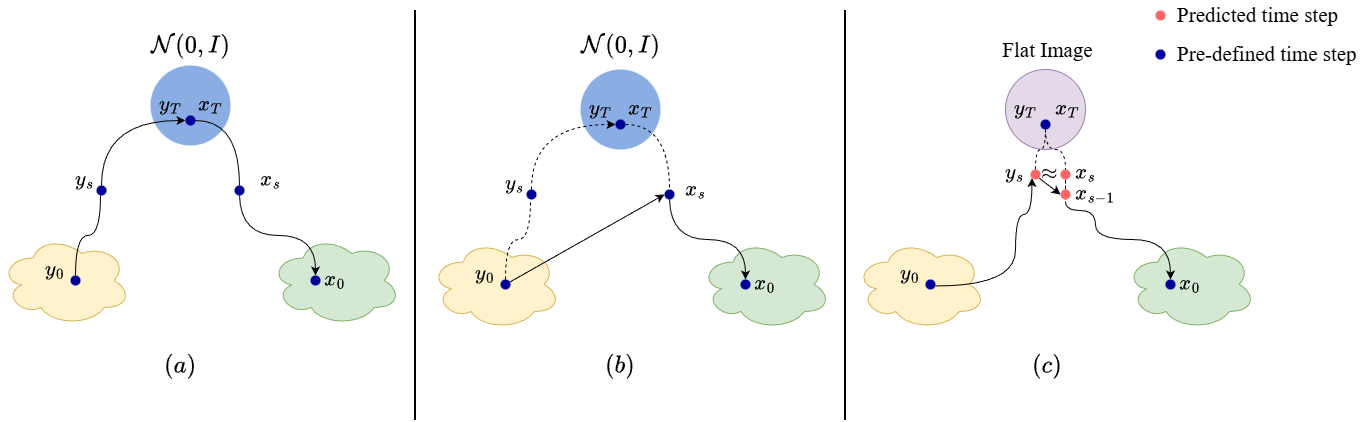}
    
    \caption{Image restoration by diffusion inversion methods. $(a)$ In inversion-based methods, the input image $y_0$ is first mapped to the noise space by the forward process. Then the extracted noise is used to solve the reverse process. $(b)$ \citep{Yue2024ArbitrarystepsIS} perturbs the model with additive noise, aligning $y_0$ to an intermediate $x_s$. This method is feasible with a vast number of reverse problems whose input is only slightly deviated from the expected output. $(c)$ Our method aims to find the optimal time step $s$, determined by the degree of degradation, at which $y_s$ and $x_s$ are highly similar. A reverse network is trained to revert $y_s \approx x_s$ to $x_0$, adapting to multiple severities.}
    \label{fig:3modelsdiff}
  
\end{figure*}

Selecting a predefined intermediate time step as the initial time step for diffusion inversion was first explored in \citep{Yue2024ArbitrarystepsIS}. However, they use a rigid noise injection strategy. While this strategy is reasonable for super-resolution tasks with a fixed scale factor—the primary focus of \citep{Yue2024ArbitrarystepsIS}—it becomes restrictive for image restoration tasks such as deblurring or dehazing or low-light enhancement, which are the main focus of our work, where the appropriate noise level depends on the severity of degradation. 

\begin{figure}[t]
  \centering

    \includegraphics[width=\columnwidth]{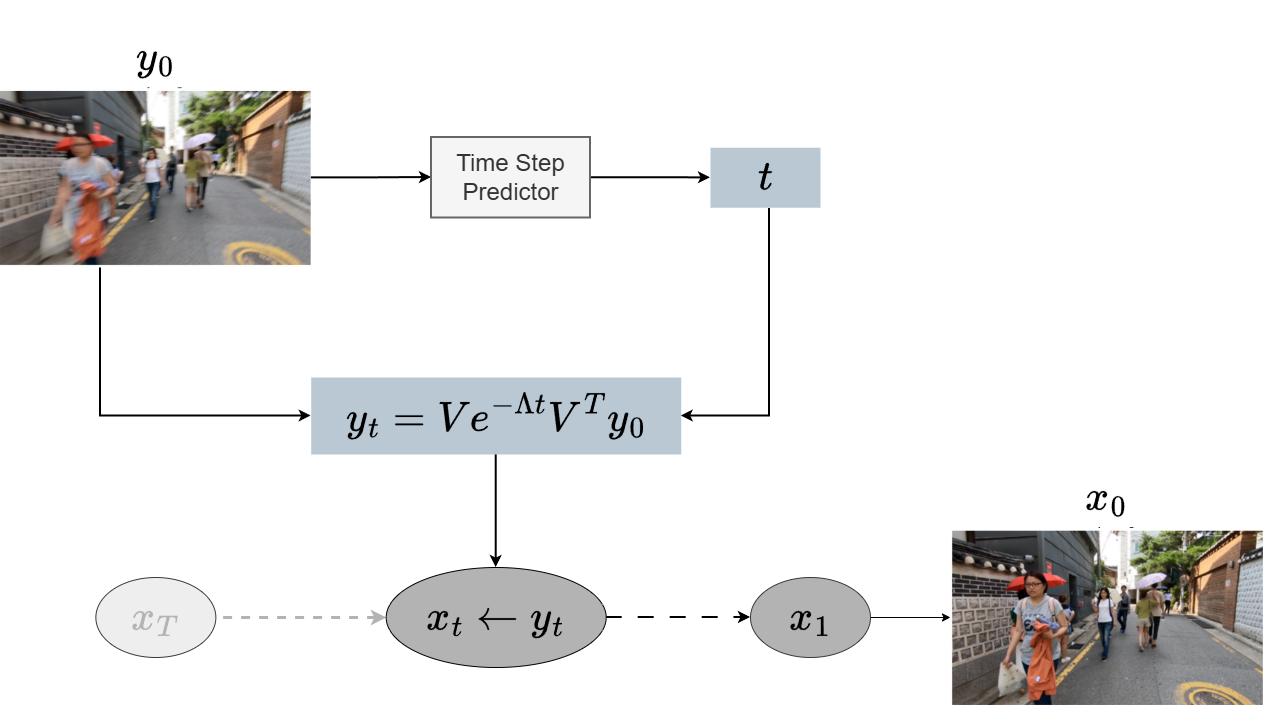}

    \caption{Inference flow of our proposed method.}
    \label{fig:timestepmodel}
\end{figure}

Addressing the limitations of fixed-step inversion within diffusion-based frameworks, we introduce a dissipation-based inversion approach that predicts the optimal initial time step for sampling. As illustrated in Fig.~\ref{fig:timestepmodel}, this learned time step predictor allows the inversion to begin at the point along the dissipation trajectory where the degraded and clean feature distributions are most aligned.


\subsection{Preliminaries}

Instead of using increasing Gaussian noise like  \citep{NEURIPS2020_4c5bcfec}, IHDM \citep{rissanen2023generative} destroys image information with heat dissipation. They propose to solve the partial differential equation for heat dissipation
\begin{equation}
\label{equa:heat_dissipation}
    \frac{\partial}{\partial t}z(i,j,t) = \Delta z(i,j,t),
\end{equation}
where $z: \mathbb{R}^2\times \mathbb{R}_+ \rightarrow \mathbb{R}$ is the idealized, continuous 2D plane of one channel of the image and $\Delta$ is the Laplace operator, can be solved by a diagonal matrix in the frequency domain of the cosine transform. Letting $z_t$ denote the solution to $\Delta$ at time step $t$, this can be efficiently computed by
\begin{equation}
\label{equa:dct}
z_t = A(t)z_0 = VD(t)V^Tz_0 = V\exp(-\Lambda t)V^Tz_0,
\end{equation}
where $V^{T}$ and $V$ are the Discrete Cosine Transform (DCT) and the inverse. The diagonal matrix $D(t)$ is the exponent of the weighting matrix for frequencies $\Lambda$ and the time step $t$, so that $D(t) = \exp(-\Lambda t)$. The intermediate diffusion state $z_t$ is constructed by gradually blurring the initial data point $z_0$ with heat dissipation. In the reverse process, \citep{rissanen2023generative} trains a generative model to invert the heat dissipation process

\begin{equation}
\label{eq:reverse_hp}
    q(z_{t-1}|z_{t}) = \mathcal{N}(z_{t-1}|f_{\theta}(z_{t},t),\sigma^2 \textit{\textbf{I}}),
\end{equation}
where the mean of $z_{t-1}$ is estimated through a neural network $f_\theta$ and has a fixed scalar variance $\sigma^2$. Similarly to DDPMs, IHDM is learned by forwarding an input image $z_0$ to a random time step $t$: $z_t = A(t)z_0$, \textit{} then minimizing the square reconstruction error between the model $f_\theta(z_t,t)$ and a ground truth target, which in this case is given by $\mathbb{E}(z_{t-1}|z_0)=A(t-1)z_0$, yielding the training loss 
\begin{equation}
    \mathbb{L}=\mathbb{E}_{t\sim \mathbb{U}(1,...,T)}\mathbb{E}_{z_t=A(t)z_0}[||A(t-1)z_0 - f_\theta(z_t,t)||^2].
\end{equation}

The notable point is that in Eq.\ \ref{equa:heat_dissipation}, $A_t$ is differentiable with respect to $t$. Thus, it is feasible to train an end-to-end pipeline including a Time Step Predictor implemented on top of the dissipation model.

\subsection{Diffusion Inversion}

\subsubsection{Inversion Trajectory}

 Fig.~\ref{fig:3modelsdiff}$(c)$ depicts the overall pipeline of our proposed method. Our goal is to restore a high-quality image $x_0$ from a given degraded image $y_0$ with the dissipation inversion. We first define the inversion trajectory as
\begin{equation}
    x_{t-1} = f_\theta(x_t,t) + \epsilon, \epsilon \sim \mathcal{N}(0,\sigma^2\textit{\textbf{I}}).
\end{equation}

The key to initiating this inversion trajectory is constructing the starting state $x_s$ $(0\leq s \leq T)$. Following Eq.~\ref{equa:heat_dissipation}, $x_s$ can be computed as

\begin{equation}
    x_s = A(s)x_0.
\end{equation}
    
 Inspired by \citep{Yue2024ArbitrarystepsIS}, we attempt to construct the starting state $x_s$ from low-quality input $y_0$. We aim to perturb $y_0$ with heat dissipation so that it becomes as similar to $x_s$ as possible. 


To effectively align $x_0$ and $y_0$ through perturbation, using a fixed predefined time step is suboptimal, as different degradation levels demand different amounts of perturbation. Consequently, the appropriate time step should adapt to the severity of degradation present in the input image $y_0$. To capture this dependency, we adopt the model of \citep{Zhang2018LearningTU} as a time-step predictor, which takes $y_0$ as input and outputs a suitable time step corresponding to its degradation level (detailed implementation can be found in the supplementary material). The intermediate state $x_s$ is then obtained by propagating $y_0$ forward to this predicted time step
\begin{equation}
\label{eq:intermediate_esimation}
    x_s \approx y_s= A(g_\phi(y_0))y_0.
\end{equation}

\subsubsection{Mutual Information}
Mutual information quantifies the statistical dependence between two images, capturing not only pixel intensities but also spatial relationships. We leverage this property to evaluate the suitability of a given time step by assessing the similarity between the corresponding intermediate distributions. Given time step $t$, $x$ and $y$ are the backward feature and the forward feature, respectively. The mutual information between $x$ and $y$ is given by
\begin{equation}
    \label{eq:mi}
    I(x,y) = \sum_{i_x,j_y}p_{x,y}(i_x,j_y)\log_2(\frac{p_{x,y}(i_x,j_y)}{p_x(i_x)p_y(j_y)}),
\end{equation}
where $i_x,j_y$ represent pixel intensity levels in $x$ and $y$, $p_{x,y}(i_x,j_y)$ indicates the joint distribution where pixels are $i_x$ and $j_y$ in $x$ and $y$. $p_x(i_x)$ and $p_y(i_y)$ are marginal distributions of the intensities of the pixels in images. The intensity level in Eq. \ref{eq:mi} is calculated with a soft histogram, where each sample contributes fractionally to multiple bins via a Gaussian. In our paper, we follow \cite{Peeples2024HistogramLF} and set the number of bins to $b=16$ and $\sigma_{mi}=\frac{0.5}{b}$ for the Gaussian.

During the training phase, we aim to maximize the mutual information at time step $t$ predicted by $g_\phi$. The loss function is defined as
\begin{equation}
    L_{I}(x_0,y_0) = -I(A(s)x_0,A(s)y_0),s=g_\phi(y_0).
\end{equation}

\subsubsection{Model Training}

\begin{algorithm}[b]
\caption{Inference}\label{alg:training}
\begin{algorithmic}
\Require Degraded image $y_0$, time step predictor $g_\phi$, diffusion model $f_\theta$, forward function $A$

\State $s \gets g_\phi(y_0)$
\State $x_s \gets A(s)y_0$

\While{$s \geq 1$}

\State $\epsilon\sim\mathcal{N}(0,\sigma^2\textbf{I})$
\State $x_s \gets f_\theta(x_s,s) + \epsilon$
\State $s \gets s-1$
\EndWhile

\State \textbf{return} $x_s$
\end{algorithmic}
\end{algorithm}


As defined in Eq. \ref{eq:intermediate_esimation}, we can easily obtain $x_s$ by forwarding $y_0$ to the predicted time step $s$. We now start the reverse process from $s$
\begin{equation}
    \hat{x}_{k-1} =  f_\theta(x_{k},k) +\epsilon, \epsilon \sim \mathcal{N}(0,\sigma^2\textbf{I}),0<k\leq s.
\end{equation}
Thus, our training objective is to minimize the distance between $\hat{x}_{k-1}$ and a ground truth target,
\begin{equation}
    L_D(x_0,y_0) = \sum_{k=0}^{s}||A(k-1)x_0 - \hat{x}_{k-1}||^2.
\end{equation}
Unlike \citep{Yue2024ArbitrarystepsIS}, which constructs a surrogate feature of $x_s$ from $y_0$, our model does not have a hard constraint to force the equality of $x_s$ and $y_s$. Hence, there is an inevitable difference between $x_s$ and $y_s$, making fully relying on the pre-trained dissipation model infeasible. Indeed, based on our experiments in \ref{subsec:ablation_study}, using a predefined model leads to poorer performance. We attempt to jointly train $f_\theta(\cdot,\cdot)$ and $g_\phi(\cdot)$ with the loss function defined as
\begin{equation}
    L = L_D(x_0,y_0) + \gamma L_I(x_0,y_0),
\end{equation}
where $\gamma=0.05$. For clarity, a detailed inference procedure is provided in Alg. \ref{alg:training}.

\subsubsection{Absence of Noise in Forward Process} In this section, we will discuss our choice of a noise-free forward process.

The key insight of this work is to align the forward-blurred image and the backward-reconstructed clean image by identifying an optimal time step at which their intermediate representations are most similar. Rather than modifying or manipulating the diffusion dynamics, our goal is to locate the time step along the predefined diffusion trajectory where the alignment between forward and backward states is maximized. Consequently, the choice of diffusion process becomes fundamental to our framework. In conventional diffusion models, the forward process progressively corrupts the original signal with Gaussian noise. While suitable for noise-like degradations, this mechanism is inherently mismatched with blur-related degradations, such as blur or haze. As a result, the intermediate states generated by Gaussian diffusion may not align well with those required for blur restoration. In contrast, the IHDM employs a deterministic forward process that can be interpreted as progressively diffusing the image via a blur-like operation. This makes it naturally compatible with blur-based restoration tasks. Empirically, in our experiment \ref{subsec:ablation_study}, we observe that at the optimal time step $t$, the mutual information $I(x_t,y_t)$ obtained under IHDM is higher than that achieved using continuous diffusion models, indicating stronger alignment between forward and backward representations.
Blurring Diffusion Model (BDM) \citep{hoogeboom2023blurring} can rigorously inject noise into the forward process of the IHDM, but the randomness injected at each forward step may excessively disrupt the signal content due to the dual nature of signal degradation. Besides, it can lead the time step predictor to sub-optimal solutions, which is a critical failure in the proposed method. Therefore, we are interested in \emph{isolating the uncertainty entirely to the reverse process}. This is particularly useful in an image restoration setting, as the forward (degradation) process is only applied midway to retain image semantic structures as much as possible. The reverse process must then be stochastic to exactly capture the uncertainty in the restoration. In fact, based on our experiments in \ref{subsec:ablation_study}, BDM produces less meaningful starting points than IHDM, leading to poorer performance in the restoration process.

\section{Experimental Results}

\begin{table*}[]
\centering
\resizebox{\linewidth}{!}{
\begin{tabular}{l|ccccc|ccccc|ccccc|c}\hline
\multicolumn{1}{l|}{\multirow{2}{*}{\textbf{Models}}} & \multicolumn{5}{c|}{Low-Light}        & \multicolumn{5}{c|}{Haze}            & \multicolumn{5}{c|}{Blur}             & \#Param \\
\multicolumn{1}{c|}{}                        & PSNR$\uparrow$& SSIM$\uparrow$  & LPIPS$\downarrow$ & FID$\downarrow$   & KID$\downarrow$   & PSNR$\uparrow$  & SSIM$\uparrow$  & LPIPS$\downarrow$ & FID$\downarrow$   & KID$\downarrow$  & PSNR$\uparrow$  & SSIM$\uparrow$  & LPIPS$\downarrow$ & FID$\downarrow$   & KID$\downarrow$   & M       \\ \hline
InvSR                                        & 18.48 & 0.809 & 0.159 & 35.91 & 10.56 & 29.36 & 0.930  & 0.041 & 39.27 & 5.95 & 28.02 & 0.799 & \textcolor{orange}{0.049} & 57.36 & 18.87 & 33.8    \\
DFPIR                                        & 18.74 & 0.843 & 0.146 & 35.12 & 10.92 & 29.63 & 0.932 & 0.043 & 30.17 & 3.22 & 28.17 & 0.813 & 0.055 & \textcolor{orange}{56.26} & \textcolor{orange}{14.44} & 31.1    \\
MoCE-IR                                      & \textcolor{blue}{20.04} & \textcolor{blue}{0.889} & \textcolor{blue}{0.124} & \textcolor{orange}{34.43} & \textcolor{orange}{9.32}  & 30.16 & \textcolor{blue}{0.978} & \textcolor{blue}{0.031} & \textcolor{orange}{29.01} & \textcolor{red}{1.92} & \textcolor{blue}{28.64} & \textcolor{orange}{0.857} & 0.051 & 56.58 & 15.37 & 25.4    \\
FoundIR                                      & \textcolor{red}{21.40} & \textcolor{red}{0.989} & \textcolor{red}{0.004} & \textcolor{blue}{34.37} & \textcolor{red}{8.92}  & \textcolor{red}{30.85}& \textcolor{orange}{0.964} & \textcolor{orange}{0.034} & \textcolor{blue}{28.98} & \textcolor{orange}{2.26} & \textcolor{orange}{28.37} & \textcolor{blue}{0.869} & \textcolor{blue}{0.045} & \textcolor{blue}{54.24} & \textcolor{blue}{13.18} & 36.3    \\ \hline
Our model                                    & \textcolor{orange}{19.98} & \textcolor{orange}{0.883} & \textcolor{orange}{0.127} & \textcolor{red}{33.92} & \textcolor{blue}{9.13}  & \textcolor{blue}{30.70} & \textcolor{red}{0.983} & \textcolor{red}{0.021} & \textcolor{red}{28.84} & \textcolor{blue}{2.02} & \textcolor{red}{29.09} & \textcolor{red}{0.880} & \textcolor{red}{0.033} & \textcolor{red}{53.99} & \textcolor{red}{13.01} & 37.7  \\ \hline
\end{tabular}
}
\caption{Quantitative comparisons to SOTAs in the all-in-one setting. KID values are scaled by a factor of 100 for readability. \textcolor{red}{Red}, \textcolor{blue}{Blue}, and \textcolor{orange}{Orange} denote the first, second, and third best, respectively.}
\label{tab:all_in_one}
\end{table*}

\begin{table*}[t]
\footnotesize
\resizebox{0.32\linewidth}{!}{
\begin{tabular}{l|ccccc}
\hline
\multirow{2}{*}{\textbf{Models}} & \multicolumn{5}{c}{\textbf{Low-light}} \\
                                 & PSNR$\uparrow$& SSIM$\uparrow$  & LPIPS$\downarrow$ & FID$\downarrow$   & KID$\downarrow$   \\ \hline
Diff LL                  & \textcolor{blue}{20.47}  & 0.879 & 0.143 & 37.36 & 13.31 \\
PNP                              & 13.83  & 0.649 & 0.492 & 60.06 & 28.49 \\
InstructP2P                      & 14.33  & 0.757 & 0.398 & 34.31 & 17.90 \\
Diff-Plugin                      & 19.95  & \textcolor{orange}{0.888} & \textcolor{blue}{0.121} & \textcolor{orange}{33.26} & \textcolor{blue}{7.28}  \\
InvSR                            & 17.85  & 0.853 & 0.145 & 34.37 & 9.09  \\
DFPIR                            & 18.94  & 0.866 & 0.140 & 36.49 & 9.27  \\
MoCE-IR                          & \textcolor{orange}{20.14}  & \textcolor{red}{0.902} & \textcolor{orange}{0.135}{} & \textcolor{blue}{32.92} & \textcolor{red}{6.15}  \\ \hline
Our model                            & \textcolor{red}{20.79}  & \textcolor{blue}{0.899} & \textcolor{red}{0.120} & \textcolor{red}{32.90} & \textcolor{orange}{8.76}  \\ \hline
\end{tabular}}
\hfill
\resizebox{0.32\linewidth}{!}{
\begin{tabular}{l|ccccc}
\hline
\multirow{2}{*}{\textbf{Models}} & \multicolumn{5}{c}{\textbf{Haze}}     \\
                                 & PSNR$\uparrow$& SSIM$\uparrow$  & LPIPS$\downarrow$ & FID$\downarrow$   & KID$\downarrow$   \\ \hline
DEA-Net                          & 30.08 & 0.961 & 0.064 & 28.90 & 3.94  \\
PNP                              & 13.68 & 0.590 & 0.307 & 57.16 & 10.28 \\
InstructP2P                      & 23.95 & 0.754 & 0.154 & 34.31 & \textcolor{blue}{1.79}  \\
Diff-Plugin                      & \textcolor{red}{31.45} & \textcolor{red}{0.996} & \textcolor{red}{0.012} & \textcolor{red}{22.07} & \textcolor{red}{0.71}  \\
InvSR                            & 29.37 & 0.948 & 0.031 & 29.68 & 3.01  \\
DFPIR                            & 30.07 & 0.946 & 0.035 & 29.38 & 2.88  \\
MoCE-IR                          & \textcolor{orange}{30.24} & \textcolor{orange}{0.984} & \textcolor{orange}{0.021} & \textcolor{orange}{28.62} & 2.67  \\ \hline
Ours                             & \textcolor{blue}{31.10} & \textcolor{blue}{0.995} & \textcolor{blue}{0.018} & \textcolor{blue}{27.96} & \textcolor{orange}{2.54}  \\ \hline
\end{tabular}}
\hfill
\resizebox{0.32\linewidth}{!}{
\begin{tabular}{l|ccccc}
\hline
\multicolumn{1}{c|}{\multirow{2}{*}{\textbf{Models}}} & \multicolumn{5}{c}{\textbf{Blur}}     \\
\multicolumn{1}{c|}{}                                 & PSNR$\uparrow$& SSIM$\uparrow$  & LPIPS$\downarrow$ & FID$\downarrow$   & KID$\downarrow$   \\ \hline
Restormer                                             & \textcolor{orange}{28.96} & \textcolor{orange}{0.879} & -     & 55.64 & 15.70 \\
PNP                                                   & 25.52 & 0.645 & 0.103 & 63.16 & 23.58 \\
InstructP2P                                           & 25.31 & 0.616 & 0.109 & 57.38 & 19.37 \\
Diff-Plugin                                           & 28.95 & 0.878 & \textcolor{blue}{0.040} & \textcolor{blue}{51.81} & 14.63 \\
InvSR                                                 & 28.44 & 0.833 & 0.048 & 52.12 & \textcolor{orange}{13.37} \\
DFPIR                                                 & 28.78 & 0.868 & 0.049 & 51.86 & 15.46 \\
MoCE-IR                                               & \textcolor{blue}{29.33} & \textcolor{blue}{0.880} & \textcolor{orange}{0.044} & \textcolor{orange}{51.84} & \textcolor{blue}{13.24} \\ \hline
Ours                                                  & \textcolor{red}{29.86} & \textcolor{red}{0.881} & \textcolor{red}{0.029}& \textcolor{red}{51.35} & \textcolor{red}{11.44} \\ \hline
\end{tabular}}

\caption{Quantitative comparison in single-task setting. From left to right: Low-light enhancement, dehazing, and debluring.}
\label{tab:quantitative_results}
\end{table*}
\begin{table}[t]
\centering
\resizebox{\columnwidth}{!}{
\begin{tabular}{c|c|c|c|c}
\hline
                Methods   & Ours  & Diff-Plugin  & InstructP2P & PNP\\ \hline                
Avg Inference Time (s) & 1.626 & 2.994  & 8.891 & 0.724    \\ \hline
\end{tabular}
}
\caption{Average inference time comparison on a single A100 GPU with the input size of $512\times512$.}
\label{tab:infer_time}
\end{table}
\subsection{Experiments Setup}
\textbf{Datasets and Metrics}
To train our models, we utilize specific datasets for each low-level task: low-light: LOL \citep{Xu2022SNRAwareLI}, dehazing: RESIDE \citep{li2019benchmarking}, and deblurring: GOPRO \citep{Nah_2017_CVPR}. We train our model in two settings: all-in-one and single-task. For testing, we evaluate on: low-light: LoLI-Street \citep{Islam_2024_ACCV}, RESIDE \citep{li2019benchmarking}, and deblurring: RealBlur \citep{Rim2020RealWorldBD}. We follow \citep{Liu2024DiffPluginRD} to employ widely adopted non-reference perceptual metrics, Fréchet Inception Distance (FID) \citep{Heusel2017GANsTB} and Kernel Inception Distance (KID) \citep{Binkowski2018DemystifyingMG}, Peak Signal-to-Noise Ratio (PSNR), Structural Similarity Index Measure (SSIM) \citep{1284395}, Learned Perceptual Image Patch Similarity (LPIPS) \citep{zhang2018perceptual}, and Deep Image Structure and Texture Similarity (DISTS) \citep{ding2020iqa}.

\noindent\textbf{Implementations} During training, we used the Adam optimizer \citep{Kingma2014AdamAM} with $\beta_{1}=0.9$ and $\beta_{2}=0.999$ and the learning rate $2e-5$. The number of epochs is set to $100$. All experiments were implemented in PyTorch.

\subsection{Comparison with State-of-the-Art Methods}
We compare the proposed methods with the state-of-the-art methods from different low-level vision tasks, including specialized models: DEA-Net \citep{chen2023dea}, Restormer \citep{Zamir2021Restormer}, Diffusion LL \citep{jiang2023low}, multi-task models DFPIR \cite{tian2025degradation}, MoCE \citep{zamfir2024complexityexperts}, and diffusion-based models: PNP \citep{Tumanyan2022PlugandPlayDF}, InstructP2P \citep{Brooks2022InstructPix2PixLT}, Diff-Plugin \citep{Liu2024DiffPluginRD}, InvSR \cite{Yue2024ArbitrarystepsIS} and FoundIR \cite{li2024foundir}. The reported results are cited directly from the papers, reproduced through released models, or retrained by us (details can be found in the supplementary material). 
\begin{figure*}[t]
  \centering

    \includegraphics[width=\linewidth]{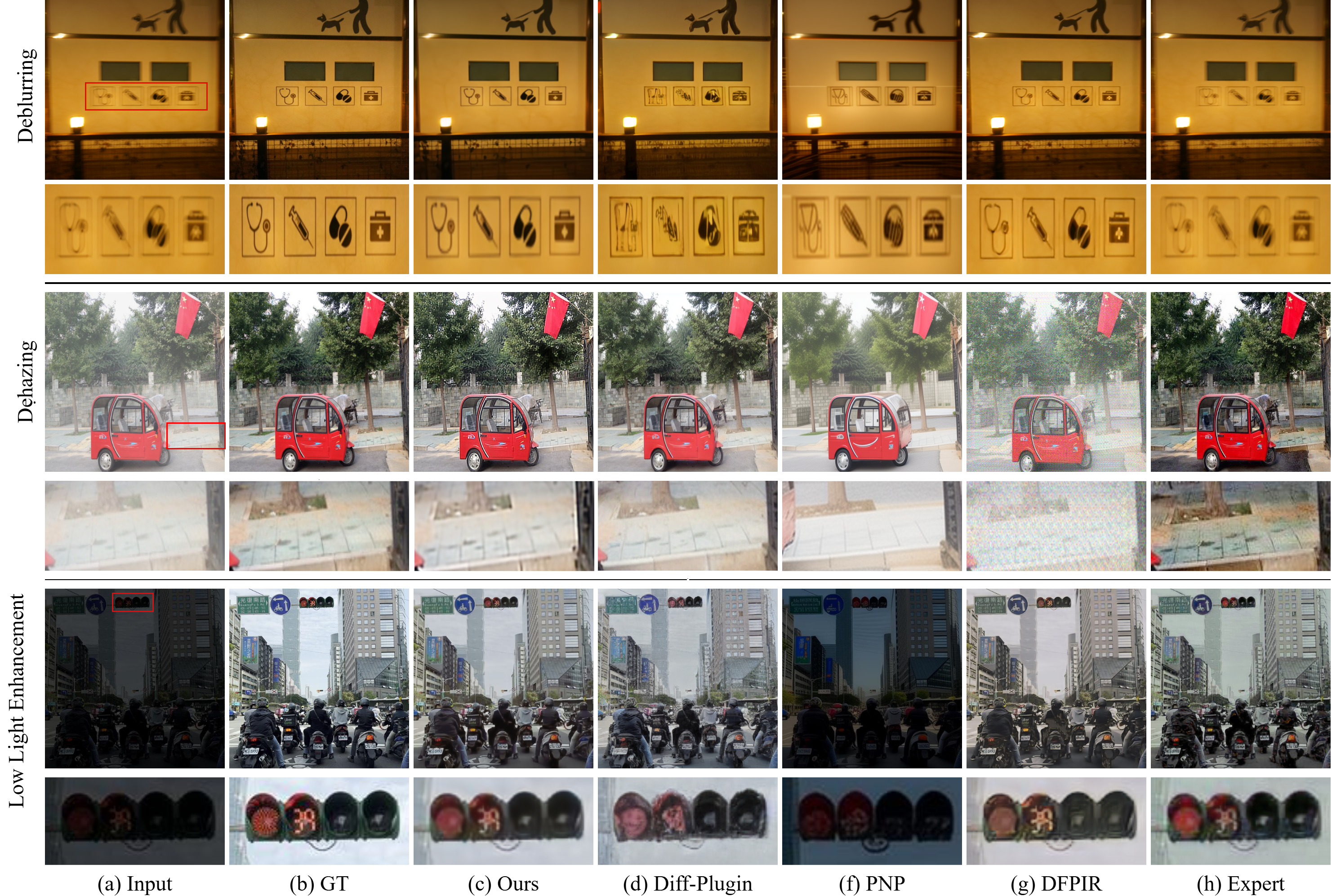}
    
    \caption{Qualitative comparison. From left to right: degraded image, ground truth, ours, Diff-Plugin \citep{Liu2024DiffPluginRD}, PNP \citep{Tumanyan2022PlugandPlayDF}, DFPIR \citep{tian2025degradation}, and expert models. Please zoom in for better visualization.}
    \label{fig:qualitative_res}
  
\end{figure*}

\noindent\textbf{Quantitative Results} In Tabs.\ \ref{tab:quantitative_results} and \ref{tab:all_in_one}, we provide the quantitative comparison of our method with other SOTA methods on test datasets. Our method achieved SOTA results overall. While Diff-Plugin is capable of producing high-quality images with low FID and KID, it often fails to reconstruct small objects like words or logos (refer to Fig. \ref{fig:qualitative_res}). Compared to other multi-task methods, our method proves to be more efficient in Deblurring and Dehazing tasks. Compared to expert models, our method delivers competitive performance.

 We also report the average inference time of our model and other diffusion-based methods running on $100$ input images in Tab.\ \ref{tab:infer_time}. The average NFE of our method is $33.83$, and the standard deviation is $9.44$. For each input image, we use the time step predicted by our model for all of the methods. The result shows that our method is notably fast within the diffusion family (Diff-Plugin).
 \citep{Liu2024DiffPluginRD}).

\noindent\textbf{Qualitative Results}  We have compared our method with different methods in terms of visual quality. The results are depicted in Fig.\ \ref{fig:qualitative_res}. From the figure, our method achieves a consistent performance in all tasks, producing high-fidelity images. First, using the Diff-Plugin \citep{Liu2024DiffPluginRD} does not ensure the accuracy of the details. It fails to construct small patterns like logos or words on signs (column 4).  Using PNP \citep{Tumanyan2022PlugandPlayDF} also falls short due to using initial noise from DDIM inversion. DFPIR can not totally remove degradation from severe scenarios (row 1).
We also depict the results of our model on real-world cases in Fig.\ \ref{fig:intro}. More results are reported in the supplementary material.
 \subsection{Model Analysis} \label{subsec:ablation_study}

\begin{table}[t]
\centering

\begin{tabular}{l|cc}
\hline
\multirow{2}{*}{Model selection} & \multicolumn{2}{c}{MI $\uparrow$}                     \\ 
                       & \multicolumn{1}{c}{RealBlur} & LOLI-Street \\ \hline
DM \citep{Song2020ScoreBasedGM}                    & \multicolumn{1}{c}{0.915}    & 0.257       \\ 
BDM \citep{hoogeboom2023blurring}            & \multicolumn{1}{c}{0.921}    & 0.325       \\ \hline
IHDM \cite{rissanen2023generative}                  & \multicolumn{1}{c}{\textcolor{red}{1.021}}    & \textcolor{red}{0.327}       \\ \hline
\end{tabular}
\caption{Mutual Information (MI) between $x_t$ and $y_t$ at the optimal time steps. We test our proposed method using different diffusion models as the backbone and the same optimization objective.}

  \label{tab:mi_differentmodels}
\end{table}
\noindent\textbf{The Selection of Models} Tab. \ref{tab:mi_differentmodels} reports the mutual information (MI) between $x_t$ and $y_t$ under different diffusion models on the RealBlur and LOLI-Street datasets. Compared to the continuous diffusion model (DM) \citep{Song2020ScoreBasedGM}, IHDM improves MI by 0.106 on RealBlur and 0.07 on LOLI-Street, indicating stronger alignment between forward and backward intermediate states. Although the Blurring Diffusion Model \citep{hoogeboom2023blurring} already improves MI over DM, IHDM provides further gains, suggesting that its deterministic heat-dissipation process better preserves structural dependencies relevant to blur restoration.

We also present the image restoration task results of the BDM and IHDM models in Tab.\ 
\ref{tab:noise_vs_nonoise}. IHDM consistently outperforms the BDM across both datasets and metrics. These results suggest that the heat-dissipation process in IHDM produces more realistic and statistically consistent restorations compared to the blurring diffusion formulation.
\begin{table}[t]
\centering
\begin{tabular}{l|cc|cc}
\hline
\multirow{2}{*}{Model selection}                    & \multicolumn{2}{c|}{RealBlur}       & \multicolumn{2}{l}{LOLI-Street}   \\
                                                    & FID $\downarrow$ & KID $\downarrow$ & FID$\downarrow$ & KID$\downarrow$ \\ \hline
DM  \citep{Song2020ScoreBasedGM}                                                 & 54.98            & 16.35            & 33.46           & 9.77            \\
BDM \citep{hoogeboom2023blurring}  & 55.10            & 16.18            & 33.39           & 9.42            \\ \hline
IHDM \cite{rissanen2023generative} & \textcolor{red}{51.35}            & \textcolor{red}{11.44}            & \textcolor{red}{32.90}           & \textcolor{red}{8.76}           \\ \hline
\end{tabular}

\caption{Comparison between using different models as backbone.}

  \label{tab:noise_vs_nonoise}

\end{table}




\noindent\textbf{Impact of Time Step on Image Restoration}
\begin{figure}[t]
  \centering

    \includegraphics[width=0.9\linewidth]{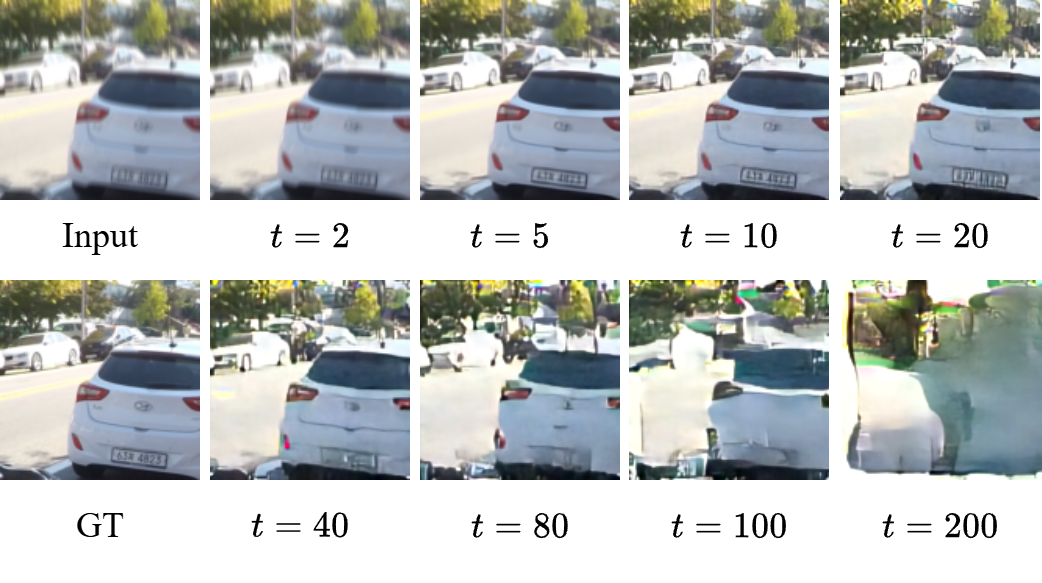}   
    \caption{Qualitative results of IHDM on GOPRO dataset \citep{Nah_2017_CVPR} at different time steps.}
  
  \label{fig:timestep_diff}
\end{figure}
\begin{figure}[t]
  \centering

    \includegraphics[width=0.9\linewidth]{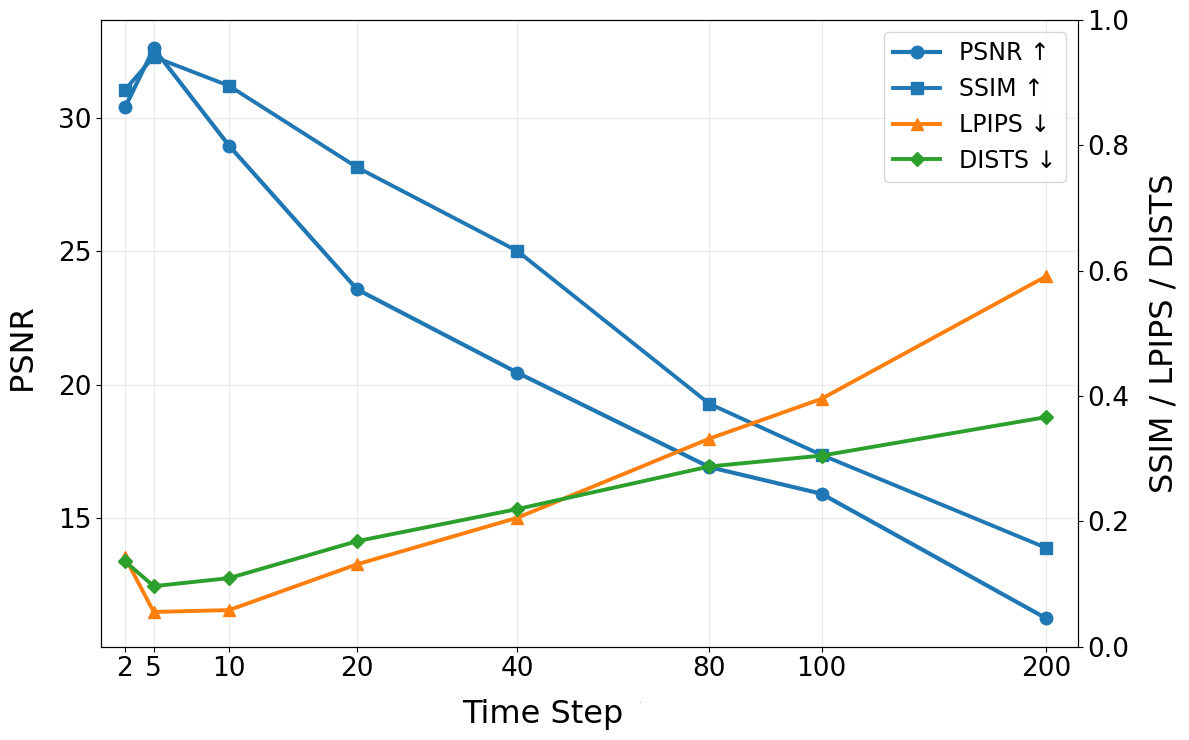}   
    \caption{Quantitative results of IHDM on GOPRO dataset \citep{Nah_2017_CVPR} at different time steps.}
  
  \label{tab:different_timestep}
\end{figure}
To illustrate how the time step can affect the restoration performance of the diffusion model, we start the inversion at multiple time steps and report the qualitative results in Fig.\ \ref{fig:timestep_diff} and the quantitative results in Fig.\ \ref{tab:different_timestep}. The experiment is conducted with the input size of $128\times128$. The model is trained by us with the setting provided in the original paper \citep{rissanen2023generative}. Based on the results, the model fails to remove degradation if we start the inversion too soon (at $t=2$). If the starting time step is sufficiently large ($\eg$ $t=5$ or $t=10$), the model can successfully revert the input back to the clean image. However, as the time step gets bigger, the structural details of the starting features get destroyed, leading to unrealistic output ($\eg$ $t=100$ or $t=200$). The trend is captured better in Fig.~\ref{tab:different_timestep}. 

 \noindent\textbf{Performance on different levels of degradation} We conducted the experiment with different levels of blur and reported the result in Tab.\ \ref{tab:different_blur_level}. The results show that our model achieves the best performance at all levels of degradation, proving better adaptability toward a range of severity.
\begin{table}[]
\centering
\resizebox{\columnwidth}{!}{
\begin{tabular}{l|ccccc}\hline
\multicolumn{1}{c|}{\multirow{2}{*}{Methods}} & \multicolumn{5}{c}{Gaussian Blur (PSNR/SSIM)}                                                                                                           \\ \cline{2-6} 
\multicolumn{1}{c|}{}                       & \multicolumn{1}{c|}{$r=0.5$}       & \multicolumn{1}{c|}{$r=1$}         & \multicolumn{1}{c|}{$r=2.5$}       & \multicolumn{1}{c|}{$r=5$}         & $r=10$        \\ \hline
InvSR                                       & \multicolumn{1}{c|}{29.88/0.889} & \multicolumn{1}{c|}{25.97/0.791} & \multicolumn{1}{c|}{22.76/0.617} & \multicolumn{1}{c|}{20.37/0.498} & 18.61/0.401 \\
DFPIR                                       & \multicolumn{1}{c|}{30.02/0.909} & \multicolumn{1}{c|}{26.36/0.818} & \multicolumn{1}{c|}{23.23/0.663} & \multicolumn{1}{c|}{21.11/0.534} & 19.18/0.464 \\ \hline
Our model                                   & \multicolumn{1}{c|}{\textcolor{red}{33.24}/\textcolor{red}{0.948}} & \multicolumn{1}{c|}{\textcolor{red}{32.04}/\textcolor{red}{0.908}} & \multicolumn{1}{c|}{\textcolor{red}{29.97}/\textcolor{red}{0.888}} & \multicolumn{1}{c|}{\textcolor{red}{26.78}/\textcolor{red}{0.814}} & \textcolor{red}{25.55}/\textcolor{red}{0.799} \\\hline
\end{tabular}}
\caption{Quantitative results on the image deblurring task with different levels of blur.  In this experiment, in order to easily modify the level of degradation, we apply a synthetic Gaussian blur, which is relatively controlled by the blur radius $r$, on 1221 test images from all categories of ImageNet.}
\label{tab:different_blur_level}
\end{table}

\begin{figure}[t]
  \centering

    \includegraphics[width=0.9\linewidth]{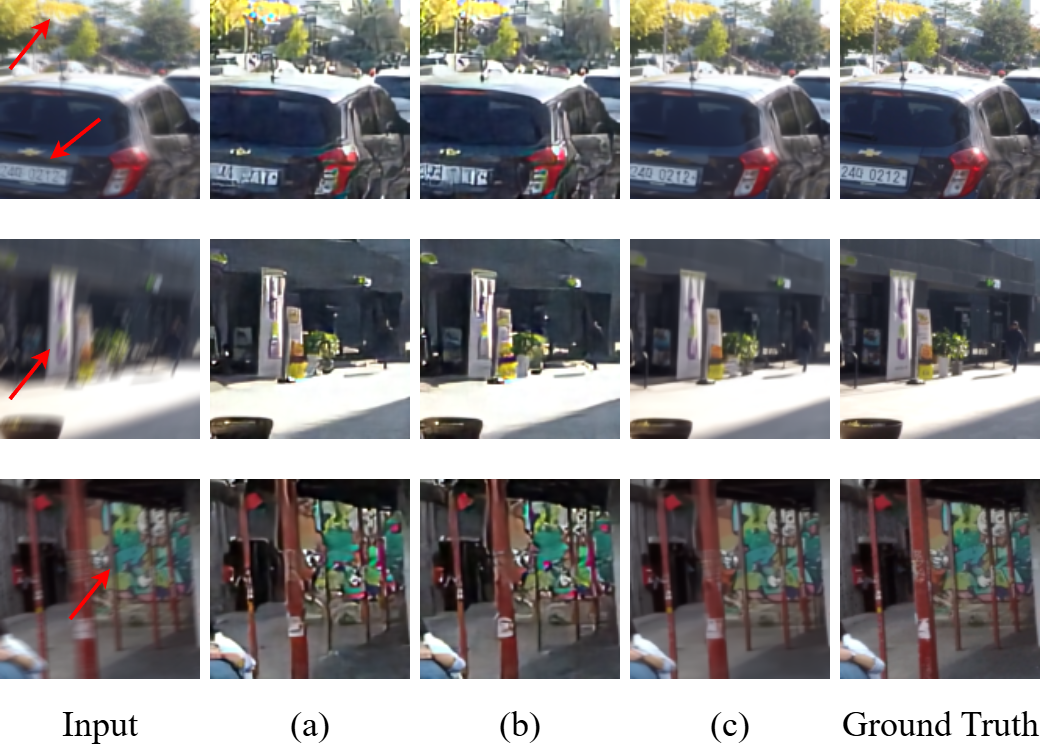}   
    \caption{Visual results on GOPRO dataset \citep{Nah_2017_CVPR}. From left to right: input, (a) pre-trained $f_\theta$ with pre-defined time steps, (b) train $g_\phi$ only, (c) train both $f_\theta$ and $g_{\phi}$, and ground truth.}
  
   \label{fig:pretrain_ablation}
\end{figure}

\noindent\textbf{The use of pre-trained model} In order to prove the effectiveness of jointly training $g_\phi$ and $f_\theta$ and the time step predictor, we compare the visual results of: (a) pre-trained IHDM with pre-defined time step, (b) pre-trained IHDM with time step predictor, and (c) our proposed models, shown in Fig.\ \ref{fig:pretrain_ablation}. The experiment is conducted with GoPro dataset \citep{Nah_2017_CVPR} and the input size is $128\times128$.

The results show a slight improvement with the second setting compared with the first one, proving that with the optimal time step $t$, the model can better restore degraded images. However, the improvement is still marginal as the model fails to adapt to the difference between $y_t$ and $x_t$. Nevertheless, the visual results produced by the pretrained model in both settings tend to be over-sharpened. On the other hand, if the models are trained together, the adaptation is better, leading to higher fidelity and quality in image generation.

\begin{figure}[t]
  \centering

    \includegraphics[width=0.9\linewidth]{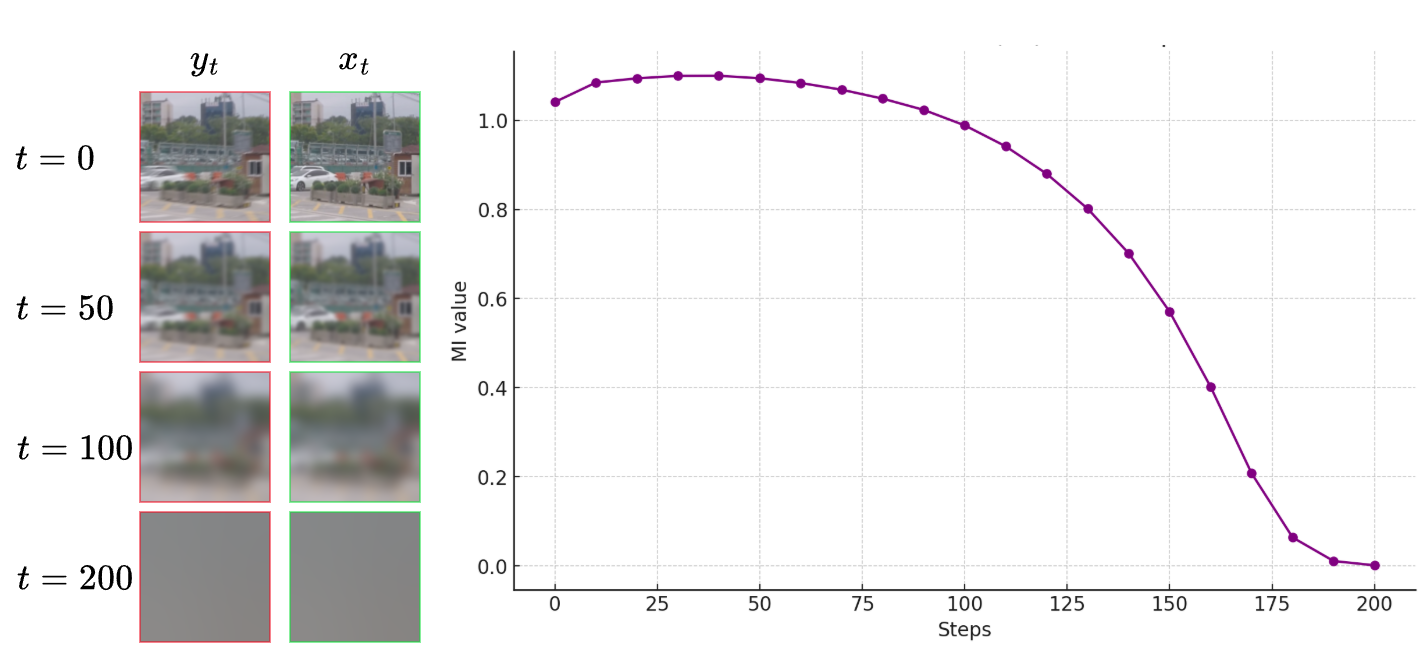}
    
    \caption{Mutual Information between $x_t$ and $y_t$ at different time steps along the diffusion process.}
  
  \label{fig:mutual_information}

\end{figure}
\noindent\textbf{Impact of Mutual Information} Fig.\ \ref{fig:mutual_information} illustrates mutual information values over time steps between $x_t$ and $y_t$. As expected, the maximum value is achieved at an intermediate state where the distributions of $x_t$ and $y_t$ are adequately overlapped. As the information destruction progresses, the MI value decays to zero despite that $x_t$ and $y_t$ become identically flat, proving that our proposed measurement determines the effectiveness of the time step not based on image distance. At time step $t=50$, although $x_t$ and $y_t$ are not perfectly aligned, the overlap can still provide meaningful information, leading to a high MI value. On the other hand, at $t=200$, $x_t$ is equal to $y_t$, but all the information is destroyed; thus, it has a low MI value.

\noindent\textbf{Ablation study with time step predictor} We conducted experiments to prove the effectiveness of our core module (TS predictor) and reported the results in Tab.\ \ref{tab:result_timestep_predictor_ab}. We have also reported the accuracy and F1 score of the time step predictor in predicting the optimal time step in Table \ \ref{tab:result_timestep_predictor}. This experiment is conducted with RealBlur-J test set. The ground truth time step, which is manually collected, is the time step with the highest mutual information of the forwarded blurry input and the forwarded counterpart. We calculate accuracy and F1 with the tolerance limit $\epsilon$, which represents the maximum distance allowed between a prediction and the ground truth for that prediction to still be considered correct. For out-of-distribution, we use random random synthetic Gaussian noise.

\begin{table}[t]
\centering

\begin{tabular}{c|c|c}\hline
     \multirow{2}{*}{Tasks}     & w/ TS Predictor & w/o TS Predictor \\
          & SSIM/LPIPS      & SSIM/LPIPS       \\ \hline
Blur & 0.880/0.033   & 0.813/0.047      \\
Haze    & 0.983/0.021     & 0.909/0.063      \\
Low-light & 0.883/0.127     & 0.789/0.178     \\\hline
\end{tabular}
\caption{Ablation study of the effectiveness of the time step predictor.}
\label{tab:result_timestep_predictor_ab}
\end{table}

\begin{table}[t]
\resizebox{\columnwidth}{!}{
\begin{tabular}{l|c|c|c}\hline
                    & $\epsilon= 0$  & $\epsilon= 1$  & $\epsilon= 3$  \\
                    & Acc $\uparrow$/ F1 $\uparrow$       & Acc $\uparrow$/ F1 $\uparrow$        & Acc $\uparrow$/ F1 $\uparrow$       \\ \hline
In Dist & 96.53/0.9823  & 98.06/0.9902  & 98.88/0.9944 \\
Out Dist & 79.39/0.8001 & 82.62/0.8311 &	85.34/0.8776  \\\hline
\end{tabular}}
\caption{Results of our time step predictor on predicting the optimal time step.}
\label{tab:result_timestep_predictor}
\end{table}

\section{Conclusion}
Our proposed methodology emphasizes the limitations of diffusion inversion with pre-defined time steps, advocating for an approach that dynamically estimates the optimal time step. By leveraging mutual information to measure the superiority of time steps, our model effectively chooses the time step that balances the overlap between source and target distributions, addressing the stochastic nature of degradation and enhancing the fidelity of image generation.

{   \small
    \bibliographystyle{ieeenat_fullname}
    \bibliography{main}
}
\end{document}